\documentclass[runningheads]{llncs}
\usepackage{graphicx}
\usepackage{epstopdf}
\usepackage{lipsum}
\usepackage{amsmath}
\usepackage{amssymb}
\newcommand{\argmin}{\operatorname{argmin}}
\newcommand{\meano}{\operatorname{mean}}
\usepackage{wrapfig}
\usepackage{algorithm}
\usepackage{algorithmic}
\usepackage{amsmath}
\usepackage{xcolor}
\usepackage{float}
\usepackage{hyperref}
\begin{document}
\title{Modelling Airway Geometry as Stock Market Data using Bayesian Changepoint Detection}
\titlerunning{Airway Geometry as Stock Market Data}
\author{Kin Quan\inst{1} \and
Ryutaro Tanno\inst{1} \and
Michael Duong\inst{2} \and \\
Arjun Nair\inst{3} \and
Rebecca Shipley\inst{4} \and
Mark Jones\inst{5} \and
Christopher Brereton\inst{5} \and
\\ John Hurst \inst{6} \and
David Hawkes\inst{1} \and
Joseph Jacob\inst{1}}

\institute{Centre for Medical Image Computing, University College London, UK \\\email{kin.quan.10@ucl.ac.uk} \and
Statistical Science, University College London, UK \and
Department of Radiology, University College Hospital, UK \and
Mechanical Engineering, University College London, UK \and
NIHR Biomedical Research Centre, University of Southampton, UK  \and
UCL Respiratory,  University College London, UK}
\authorrunning{Quan et al.}

\maketitle              % typeset the header of the contribution
\begin{abstract}
Numerous lung diseases, such as idiopathic pulmonary fibrosis (IPF), exhibit dilation of the airways. Accurate measurement of dilatation enables assessment of the progression of disease. Unfortunately the combination of image noise and airway bifurcations causes high variability in the profiles of cross-sectional areas, rendering the identification of affected regions very difficult. Here we introduce a noise-robust method for automatically detecting the location of progressive airway dilatation given two profiles of the same airway acquired at different time points. We propose a probabilistic model of abrupt relative variations between profiles and perform inference via Reversible Jump Markov Chain Monte Carlo sampling. We demonstrate the efficacy of the proposed method on two datasets; (i) images of healthy airways with simulated dilatation; (ii) pairs of real images of IPF-affected airways acquired at 1 year intervals. Our model is able to detect the starting location of airway dilatation with an accuracy of 2.5mm on simulated data. The experiments on the IPF dataset display reasonable agreement with radiologists. We can compute a relative change in airway volume that may be useful for quantifying IPF disease progression. The code is available at \url{https://github.com/quan14/Modelling_Airway_Geometry_as_Stock_Market_Data}
 
\end{abstract}

\section{Introduction}\label{ss:intro}
% In fibrosing lung disease, contraction of the lung interstitium pulls on airway walls, and this dilatation is termed traction bronchiectasis. Airway dilatation calculated using crude lobar-level visual scores have been shown to be powerful predictors of outcome in idiopathic pulmonary fibrosis (IPF) \cite{Jacob2017}. More precise and automated measures that can identify and quantify airway dilatation over time on serial computed tomography (CT) imaging would be valuable as potential endpoints for IPF drug trials as a more sensitive measure of disease progression.

In fibrosing lung disease, contraction of the lung interstitium pulls on airway walls, and causes a dilatation. Airway dilatation is calculated using crude visual scores have been shown to be powerful predictors of outcome in idiopathic pulmonary fibrosis (IPF) \cite{Jacob2017}. Computed Tomography (CT) coupled with image analysis algorithms to measure the airway lumen can provide an excellent tool to quantitatively assess these dilatation. CT gives a very high image contrast between air within the airway lumen and the surrounding airway wall. However, airway lumen measuring algorithms are subjected to sources of noise. Variations in voxel sizes \cite{Achenbach2009} and CT reconstruction algorithms \cite{Gu2013a} have been shown to affect the precision of the cross-sectional area measurements of the airways. In addition, there can be variations in airway cross-sectional area due to normal biological processes such as points of airway bifurcation where the airway transiently dilates \cite{Quan2018,Gazourian2017}. These sources of noise makes post-processing of measurements difficult to locate dilatation on an airway track. Identifying the point of dilatation and subsequently calculating the volume of abnormal airways in the lung potentially provides a sensitive means for quantifying subtle worsening of disease in IPF across longitudinal scans.

In this work, we introduce a variant of Bayesian Changepoint Detection (BCPD) models to automatically identify the location of airway dilatation in the presence of strong measurement noise, given two measurements of the same airway acquired at different time points. BCPD model, typically used in DNA sequencing \cite{Siegmund2011} and stock market data analysis \cite{Mikosch2004}, aims to capture abrupt variations in the underlying distributions of a given signal or a time series. The method processes a series of airway cross-sectional area changes between baseline (first) and follow-up (second) CT scans, and generates the posterior distribution over multiple possible points of abnormal variations, whose the mode is taken as the final prediction. We test the efficacy of the method on (1) CT images of healthy airways with simulated dilatation, and (2) pairs of real images of IPF-affected airways acquired approximately  1 year intervals. For the simulated dataset, we measured the detection accuracy with respect to the commonly used naive thresholding and maximum likelihood based methods \cite{Lavielle2005}. For the longitudinal IPF dataset, we compare the predictions of our model to measurements from radiologists based on two different protocols.

\section{Method}

% In this section, we introduce our Bayesian model used to detect progression ofidiopathic pulmonary fibrosis (IPF) using changes in airway track cross sectionalarea on two consecutive serial CT scans performed 1 year apart. The method issummarise as follows; first we acquire a series of cross-sectional measurementsof a chosen airway track from both baseline and follow-up scans. We model thechanges in area as a time series. Secondly, we applied a Bayesian changepointmodel  to  detect  abrupt  dilatation  of  the  airways.  Part  of  the  implementationrequires an application of the RJMCMC sampling framework.

In this section, we introduce our method for quantifying progression of IPF on a patients serial CT scans. The method proceeds as follows. Firstly, we fit a tubular shape model to the airways in both baseline and follow-up CT scans, and acquire estimates of the cross-sectional areas along them (Sec.~\ref{PreProMethod}). We then treat the difference in the cross-sectional areas as a 1D signal, and employ the proposed Bayesian changepoint model to estimate the posterior distribution over locations of abrupt airway dilatation (Sec.~\ref{ss:bayesianmodel} and \ref{ss:rjmc}). Lastly, we post-process this posterior distribution to determine the region of dilatation (Sec.~\ref{sec:postprocess}).

\subsection{Airway Pre-processing}\label{PreProMethod}
% The progression of pulmonary airway disease is manifested as a variation in the airway geometry (citation?). 

\begin{figure}[h]
	\centering
	\includegraphics[width=\linewidth]{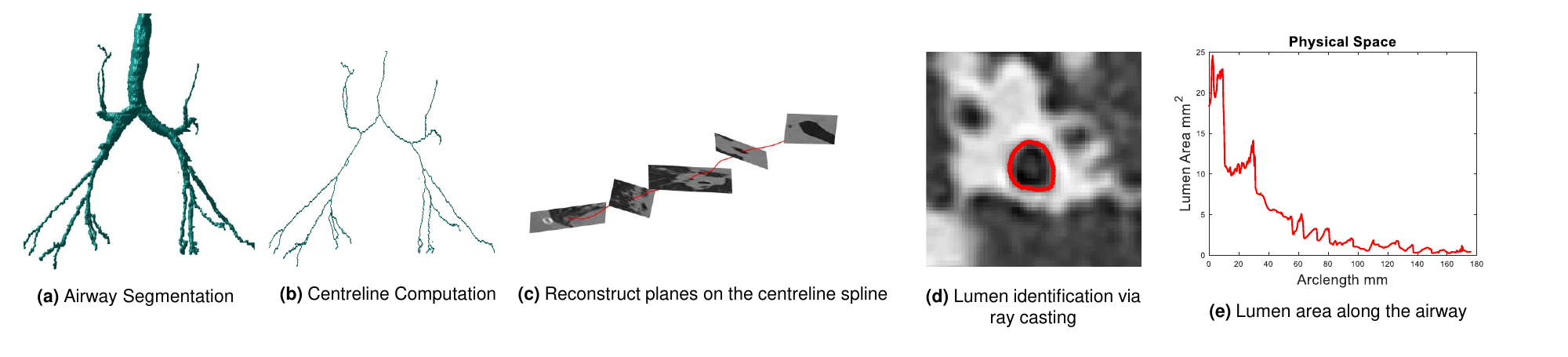}
	\caption{Summary of the pipeline developed by Quan et al \cite{Quan2018}. We have implemented the method as part of the pre-processing stage to model the geometry as a 1D signal.}
	\label{Pipeline_method_IPMI}
\end{figure}

In this work, for each airway track, we acquire a series of cross-sectional area measurements using the method proposed by Quan et al \cite{Quan2018} (Fig.~\ref{Pipeline_method_IPMI}). Following a semi-manual segmentation of the airway, the method computes the airway centreline and the corresponding normal planes at contiguous intervals, each of which is then used to estimate the cross-sectional area. The final output is a 1D function of cross-sectional area along the airways for baseline $f_{B}(x)$ and the follow-up $f_{F}(x)$ scans. 

% For a given airway track, we acquired a series of contiguous cross-sectional areameasurements  using  the  method  proposed  by  Quan  et  al  [15]  (Fig.  1).  Afterobtaining an airway segmentation and computing the airway centreline, cross-sectional planes were constructed at contiguous intervals along a smoothed cen-treline. Area measurements were taken in each of the planes. For our work, weconsider two series of the same airway on the baseline and follow-up scan.

The next task was to align the signals on both baseline $f_{B}(x)$ and follow-up $f_{F}(x)$ scans. To this end, we resample the signal to 1mm using cubic interpolation. We considered the first 50 points on both signals $ g_{B}, g_{F}$ from the start of the carina. To register the two airways together, we apply the transformation $f_{F}(x - a)$ where
\begin{equation}
a = \argmin_{a \in [-5,5]} \left|\left| \log\left( \frac{g_{B}(x)}{g_{F}(x-a)}\right) \right|\right|_{2}.
\end{equation}
The longest of the two signal were truncated from the right hand side such that both signals were of the same length. For the rest of the methodology, we will only consider the series difference defined as $\boldsymbol{y} = \log(f_{F}) - \log(f_{B})$.  %We then defined the series difference as $ \boldsymbol{y} = \log(f_{FoUp}) - \log(f_{Base})$.
%\textcolor{red}{(Ryu): the second paragarph needs to be clearer.}

\subsection{Bayesian Changepoint Model}\label{ss:bayesianmodel}
% We begin by introducing standard Bayesian methodology for detecting multiple changepoints \cite{green,STEPHENS}, using a Reversible Jump Markov Chain Monte Carlo (RJMCMC) to traverse the posterior space of changepoints. 

% We begin to introduce our changepoint detection algorithm that uses a Reversible Jump Markov Chain Monte Carlo (RJMCMC), based on the works of Green \cite{green} and Stephens \cite{STEPHENS}.

The progression of fibrosing lung diseases can manifest itself as dilatation of the internal airways. Thus, we hypothesise that at the start of dilatation, the series $\boldsymbol{y}$ undergoes an abrupt variation, which we refer to as a changepoint. More formally, given signal $\boldsymbol{y}=(y_1,\dots,y_n)$ of length $n$, we define a changepoint $\tau$ as the location where there exists a change in parameters $\theta$ in the underlying distribution $F$. In other words, at changepoint $1<\tau<n$, the observations $\boldsymbol{y}$ can be separated at $\tau$ such that:

% Here we introduce our probabilistic model of changepoints in time series data. More formally, given time series $\boldsymbol{y}=(y_1,\dots,y_n)$ of length $n$, we define a change-point $\tau$ as the location where there exists a change in parameters $\theta$ in the underlying distribution $F$. In other words, at changepoint $1<\tau<n$, the observations $\boldsymbol{y}$ can be separated at $\tau$ such that:
\begin{equation}
\boldsymbol{y} = \begin{cases}
(y_1,\dots,y_{\tau}) \sim F(\theta_1)\\ 
(y_{\tau+1},\dots,y_n) \sim F(\theta_2)
\end{cases},
\end{equation}
where $\theta_1 \neq \theta_2$. This definition can be naturally extended to the scenario with $M$ changepoints; we denote the changepoint location vector by $\tau=(\tau_1,\dots,\tau_k)$, with parameters $\theta=(\theta_1,\dots,\theta_{k+1})$ for each respective segment. For ease of notation, we also denote $\tau_0=1$ and $\tau_{k+1}=n$. Assuming statistical independence between segments, the likelihood factorises as:
\begin{equation}
p(\boldsymbol{y}|\tau,\theta,M) = \prod_{l=1}^{k+1}F(y_{\tau_{l-1}:\tau_{l}}|\theta_l),
\end{equation}
where $y_{\tau_{l-1}:\tau_{l}} = (y_{\tau_{l-1}},\dots,y_{\tau_l})$ We also specify prior distributions on the the number of changepoints $p(M;\delta)$, the locations of the changepoints $p(\tau|M; \gamma)$, and the parameters of the corresponding segments $p(\theta|M; \beta)$
% \begin{eqnarray}
% \theta &\sim& p(\beta) \notag \\
% \tau &\sim& p(\gamma) \notag\\
% M &\sim& p(\delta)
% \end{eqnarray}  
where $\beta$, $\gamma$ and $\delta$ represent the hyper-parameters. 

% The main quantities of concerns are the posterior distribution for changepoints $p(\tau|\boldsymbol{y})$ and the posterior distribution for the number of changepoints $p(M|\boldsymbol{y})$. 
% The posterior distribution of the parameters $p(\theta|\boldsymbol{y})$ will be a consequence of the methodology we employ and can be ignored in regards to providing additional information to the airways data.

Given the likelihood and the prior distributions above, we would like to estimate the posterior distributions over the number of changepoints $p(M|\boldsymbol{y})$, the locations of changepoints $p(\tau|\boldsymbol{y})$ and the parameters of the respective segments $p(\theta|\boldsymbol{y})$. Commonly in Bayesian changepoint analysis, likelihoods with a conjugate prior are chosen to allow calculation of analytical posteriors for changepoints and parameters, enabling faster convergence and better accuracy \cite{Gelman2013}. However, in our work, we wish to relax this conjugacy assumption in order to allow flexibility in the choices of prior and likelihoods. In the next section, we describe the inference scheme in such highly general setting, viable for arbitrary forms of the likelihood distribution $F$ and the prior. In this paper, $F$ was defined as the Student t distribution with parameters $\theta = (\nu, \mu, \sigma^2)$ with degrees of freedom $\nu$, mean $\mu$ and variance $\sigma^2$. We chose t-distribution for its robustness to outliers \cite{Prince2012} caused by noise from the area measurements.

%\textcolor{red}{(Ryu): Don't forget to mention $F$ is defined as student-t distribution $\theta_l = (\nu_{l}, \mu_{l}, \sigma_{l})$}. 

% (Ryu): an example why the poterior is intractable if necessary. 
% For example, the posterior over $\tau$ is given by $p(\tau|\boldsymbol{y};M) \propto p(\tau|M)p(M)\cdot \int p(\boldsymbol{y}|\theta,\tau;M) p(\theta|\tau;M)p(M) d\theta dM $ and requires integrating out $M$ and $\theta$. 

% By Bayes' rule, :
% \begin{eqnarray}
% p(\tau|\boldsymbol{y};M) &\propto& p(\boldsymbol{y}|\tau;M) p(\tau;M) \label{Bayesmarginal} \\
% &\textup{where}&\enspace \: p(\boldsymbol{y}|\tau;M)=\int_{\Theta}p(\boldsymbol{y},\theta|\tau;M) d\theta = \int_{\Theta}p(\boldsymbol{y}|\theta,\tau;M) p(\theta|\tau;M) d\theta \notag
% \end{eqnarray}

% The posterior distribution of the parameters $p(\theta|\boldsymbol{y})$ will be a consequence of the methodology we employ and can be ignored in regards to providing additional information to the airways data.

\subsection{Reversible Jump MCMC for Posterior Inference}\label{ss:rjmc}
Posterior inference with our model possesses two challenges. Firstly, without the conjugacy assumption, computing the posterior distributions is intractable. Secondly, the dimensionality of the posterior distribution over the changepoints $\tau$ is given by $M$ and varies during inference. To combat the first problem, we use the Metropolis-Hasting (MH) algorithm \cite{Chib1995}, a variant of Markov Chain Monte Carlo (MCMC) methods that can sample from the posterior, with or without conjugacy. Given that the number of changepoints $M$ is known, MH can be used to sample from the posterior distributions over the changepoints $\tau$ and segment parameters $\theta$. To address the second problem of varying posterior dimensionality $M$, we extend the above sampling scheme to the Reversible Jump MCMC framework \cite{green2009reversible}. Taken all together, the method is capable of traversing the full posterior distributions for $M, \tau, \theta$ and we refer to this as Reversible Jump Metropolis Hasting (RJMH) algorithm. 

\paragraph{\textbf{Overview of RJMH:}} The RJMH proceeds by randomly executing one of four possible moves at each iteration (see Algorithm~\ref{alg:rjmh} for the pseudo-code):
\begin{enumerate}
    \item The parameters of every segment is updated.
    \item  One of the changepoints is selected randomly and updated, and the parameters of its neighbouring two segments are updated.
    \item Randomly choose a location in the time series and add it as a changepoint. The parameters of the resultant two new segments are also sampled.
    \item One of the changepoints is randomly removed and the parameters of the new segment are sampled
\end{enumerate}

We denote these moves as: $\curlyvee_{\theta},\curlyvee_{\tau},\curlyvee_{M \rightarrow M+1},\curlyvee_{M+1 \rightarrow M}$ respectively.  We also define the maximum number of changepoints $k_{max}$ and at the boundary cases for $k$, we impose restrictions such that $\curlyvee_{M+1 \rightarrow M}$ is not possible when $k=0$  and $\curlyvee_{M \rightarrow M + 1}$ is not possible when $k=k_{max}$. Each move updates the appropriate subset of parameters $\theta, \tau$ by sampling from the corresponding proposal distributions $q(\theta_{new}|\theta_{old})$ and $q(\tau_{new}|\tau_{old})$, and is only executed if it passes the associated acceptance criteria $\alpha$. %An overview of these moves can be described as: (i) The parameters of every segment is updated. (ii) One of the changepoints is selected randomly and updated, and the parameters of its neighbouring two segments are updated. (iii) Randomly choose a location in the time series and add it as a changepoint. The parameters of the resultant two new segments are also sampled. (iv) One of the changepoints is randomly removed and the parameters of the new segment are sampled. The proposal distributions and acceptance criteria of these moves take different forms, and next we describe them in detail.

We can subdivide these four moves into two sets: fixed $M$ (MH Steps) and changing $M$ (RJ Steps) i.e. resampling parameters and moving changepoints does not affect the changepoint model $M$ whereas adding or deleting a changepoint changes the value of $M$.

\paragraph{\textbf{MH Steps:}} For $\curlyvee_{\theta}$, we set  $q(\theta_{new}|\theta_{old}) = (\mu_{old},\sigma_{old}^2,\nu_{old})+$N$(0,\epsilon)$. This step resamples parameters of each segment by proposing Gaussian perturbations around the current values of parameters for all segments. 

For $\curlyvee_{\tau}$, we set  $q(\tau_{new}|\tau_{old}) = \tau_{old}+(-1)^b$Poi$(\lambda)$ where $b\sim$Binary$[0,1]$. This step selects a changepoint $\tau$ at random and shifts it with a Poisson perturbation. The segments neighbouring this new changepoint location have parameters $\theta$ resampled as in move $\curlyvee_{\theta}$ using the current segment parameters.

Note that each proposal in the MH steps will be symmetrical and therefore, cancel out with the reverse proposal term (See Algorithm~\ref{alg:rjmh}).

\paragraph{\textbf{RJ Steps:}} For the changing $M$ moves, the algorithm moves in dimensionality and we need to introduce alternative proposal terms for the new segments produced as there are no current segment parameters to base our proposals on. 

\begin{figure}
	\includegraphics[width=\linewidth]{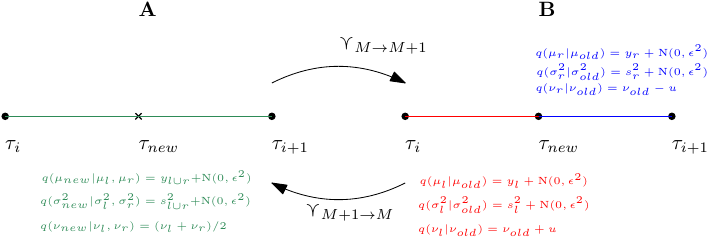}
	\caption{A schematic diagram describing the RJ steps. For $\curlyvee_{M+1 \rightarrow M}$, the changepoint $\tau_{new}$ (the central cross in state A) is removed, the parameters for the new concatenated segment $(\mu_{new}, \sigma^{2}_{new},\nu_{new})$ are updated through the green equations. For $\curlyvee_{M+1 \rightarrow M}$, the changepoint $\tau_{new}$ (the central dot in state B) is added, creating a left segment $(\mu_{l}, \sigma^{2}_{l},\nu_{l})$ and right segment $(\mu_{r}, \sigma^{2}_{r},\nu_{r})$ which are updated through the red and blue equations respectively. Note that $u \sim \text{N}(0,\epsilon^{2})$ and $y_{i},s^2_{i}$ are the mean and variance respectively of data within the coloured segment.}
	\label{Birth_and_Death_moves}
\end{figure}

For $\curlyvee_{M \rightarrow M+1}$ i.e. $M_{new} = M + 1$, we proposed random new changepoints over our data, $\tau_{new}\sim$U$[1,n-1] $. The proposed $\tau_{new}$ split an existing segment into a new left segment $\theta_{l} = (\mu_{l},\sigma^2_{l},\nu_{l})$ and new right segment $\theta_{r} = (\mu_{r},\sigma^2_{r},\nu_{r})$. Our proposal for $\mu_{i},\sigma^2_{i}$ are defined by a Gaussian perturbation on empirical values of the respective $i = l,r$ segments (Fig. \ref{Birth_and_Death_moves}). The proposal for $\nu_{i}$, is Gaussian perturbation of the previous update $\nu_{old}$. Due to dependence of the $\nu_{i}$ proposal, a Jacobian term is introduced $|J_{M \rightarrow M+1}| = 2$. 

For $\curlyvee_{M+1 \rightarrow M}$ i.e. $M_{new} = M - 1$, we remove a changepoint $\tau_{new}$. As before, proposals for $\mu_{new}$, $\sigma^2_{new}$ are defined using empirical values of the segments and Gaussian perturbation (Fig. \ref{Birth_and_Death_moves}). The proposal for $\nu_{new}$ is the mean of the previous $\nu_{l},\nu_{r}$. The move introduces Jacobian term $|J_{M+1 \rightarrow M}|  =0.5$.

\paragraph{\textbf{Initialization:}} The overall algorithm structure can be seen in the pseudo-code Algorithm~\ref{alg:rjmh}. With the model defined, we mentioned the priors used and address some common implementation techniques. The priors used are given as follows:
\begin{align}
\mu &\sim \textup{N}(0,1),\\ 
\sigma^2 &\sim \textup{Scaled-}\textup{inv-}\chi^2\left(5,0.4^2\right),\\
\nu &\sim \textup{U}[2,100],\\
M &\sim\textup{Bin}\left(n-1,\,\frac{0.5}{n-1}\right).
\end{align}
The hyper-parameters for $\mu,\sigma^2,\nu$ were chosen to be non-informative and within plausible ranges. In order to reduce the number of changepoints we could detect (in the case of acquisition noise from the CT), we set the expectation for $M$ to be sufficiently low. In terms of implementation, we follow a standard procedure of setting a burn-in for the number of iterations to ignore any potential issues with intialisation at approximately 25\% of the total iteration count. To remove any chance of auto-correlation, we thin the number of samples by only storing the $5^{\textup{th}}$ iteration, after the burn-in period. Additionally, we choose the number of iterations, $N$, to be sufficiently large in order to achieve convergence.

	\begin{algorithm}
		\caption{RJMH Changepoint Detection \cite{green2009reversible}.}
		\label{alg:generalised_em}
		\scriptsize
		\begin{algorithmic}
			%\Require Topology $\mathbb{T}$, parameters $\theta$
			\STATE \textbf{Data}: Single airway series, $\textbf{y}=(y_1,\dots,y_n)$
			\STATE \textbf{Model Parameters}: $M$ = \# changepoints, $\theta$ = segment parameters, $\tau $= changepoint location
			\STATE \textbf{RJMCMC Parameters}: $N$ = iterations, $B$ = burn-in, $T$ = Thinning
			\STATE \textbf{Initialisation}: Set $s = (M, \theta, \tau)$
			\STATE \textbf{Initialise set $s$ with no changepoints and empirical segment parameters}
			
			%%%%%%%%% (Ryu) reformatting
		\FOR{$N$ iterations}
		\STATE \textbf{Sample} $\curlyvee_{i} \sim \text{Uniform}\{\curlyvee_{\theta},\curlyvee_{\tau},\curlyvee_{M \rightarrow M+1},\curlyvee_{M+1 \rightarrow M}$\} 
		\IF{$\curlyvee_{i} =  \curlyvee_{\theta}$} 
		\STATE \textbf{MH-step (1): Resample Segment Parameters}
		\STATE Resample $\theta_{new} \sim q(\theta_{new}|\theta)$ for each segment
		\STATE Compute acceptance ratio: 
		$ \alpha = \textup{min}\Bigg\{1,\frac{p(\textbf{y}|\tau,\theta_{new},M)} {p(\textbf{y}|\tau,\theta,M)}\frac{p(\theta_{new})}{p(\theta)}\frac{q(\theta|\theta_{new})}{q(\theta_{new}|\theta)}\Bigg\}$
		\STATE Generate $\beta\sim \text{Uniform}[0,1]$
		\IF{$\beta \geq \alpha$}
		\STATE  Accept new state
		\STATE $s=(M,\theta_{new},\tau)$ 
		\ENDIF
		\ELSIF{$\curlyvee_{i} =  \curlyvee_{\tau}$}
		\STATE \textbf{MH-step (2): Move changepoint}
		\STATE Move a changepoint $\tau_{new} \sim q(\tau_{new}|\tau)$ and propose new segment parameters $\theta_{new} \sim q({\theta_{new}}|\theta)$
		\STATE Compute acceptance ratio:
		$\alpha = \textup{min}\Bigg\{1,\frac{p(\textbf{y}|{\tau_{new}},{\theta_{new}},M)} {p(\textbf{y}|\tau,\theta,M)}\frac{p({\theta_{new}})}{p(\theta)}\frac{q(\tau|{\tau_{new}})}{q({\tau_{new}}|\tau)}\frac{q(\theta|{\theta_{new}})}{q({\theta_{new}}|\theta)}\Bigg\}$
		\STATE Generate $\beta\sim \text{Uniform}[0,1]$
		\IF{$\beta \geq \alpha$}
		\STATE  Accept new state
		\STATE $s=(M,{\theta_{new}},{\tau_{new}})$
		\ENDIF
		\ELSIF{$\curlyvee_{i} =  \curlyvee_{M \rightarrow M+1}$}
		\STATE \textbf{RJ-step (3): Add changepoint}
		\STATE Propose new changepoint $M_{new}$ as ${\tau_{new}}$ and propose new segment parameters ${\theta_{new}}$
		\STATE Compute acceptance ratio: 
		\begin{equation*}
        \begin{split}
		\alpha= \textup{min}\Bigg\{1,\frac{p(\textbf{y}|{\tau_{new}},{\theta_{new}},{M_{new}})} {p(\textbf{y}|\tau,\theta,M)}\frac{p({\tau_{new}})p({\theta_{new}})p({M_{new}})}{p(\tau)p(\theta)p(M)} \\
		\frac{q(\tau|{\tau_{new}})}{q({\tau_{new}}|\tau)}\frac{q(\theta|{\theta_{new}})}{q({\theta_{new}}|\theta)}\frac{q(M|{M_{new}})}{q({M_{new}}|M)} \, |J_{M \rightarrow M+1}| \Bigg\} 
		\end{split}
        \end{equation*}
		where $|J_{M \rightarrow M+1}|=2$ and $q({M_{new}}|M)$ is symmetrical to $q(M|{M}_{new})$.
		\STATE Generate $v\sim \text{Uniform}[0,1]$
		\IF{$v \geq \alpha$}
		\STATE  Accept new state
		\STATE $s=({M_{new}},{\theta_{new}},{\tau_{new}})$ 
		\ENDIF
		\ELSIF{$\curlyvee_{i} =  \curlyvee_{M+1 \rightarrow M}$}
		\STATE \textbf{RJ-step (4): Delete changepoint}
		\STATE Delete a changepoint $M_{new}$ as $\tau_{new}$ and propose new segment parameters ${\theta_{new}}$
		\STATE Compute acceptance ratio 
		\begin{equation*}
        \begin{split}
        \alpha = \textup{min}\Bigg\{1,\frac{p(\textbf{y}|\tau{new},{\theta_{new}},{M_{new}})}{p(\textbf{y}|\tau,\theta,M)}\frac{p(\tau_{new})p({\theta_{new}})p({M_{new}})}{p(\tau)p(\theta)p(M)}\\
        \frac{q(\tau|\tau_{new})}{q(\tau_{new}|\tau)}\frac{q(\theta|{\theta_{new}})}{q({\theta_{new}}|\theta)} \frac{q(M|{M_{new}})}{q({M_{new}}|M)} \, |J_{M+1 \rightarrow M}| \Bigg\}
        \end{split}
        \end{equation*}
		where $|J_{M+1 \rightarrow M}|=\frac{1}{2}$ and $q({M_{new}}|M)$ is symmetrical to $q(M|{M_{new}})$.
		\STATE Generate $v\sim \text{Uniform}[0,1]$
		\IF{$v \geq \alpha$}
		\STATE  Accept new state
		\STATE $s=({M_{new}},{\theta_{new}},\tau_{new})$
		\ENDIF
		\ENDIF
		\ENDFOR
		\STATE \textbf{Remove first $B$ samples and keep every $T^{th}$ sample}
		\STATE \textbf{Return:} $\{s\}^{T}_{t=1}$
            \STATE $\,\,$
    	\end{algorithmic}
	\label{alg:rjmh}
\end{algorithm}

\subsection{Locating airway dilatation} \label{sec:postprocess}
% \begin{figure}[h]
% 	\centering
% 	\includegraphics[trim={0 0 0 0},clip,width=0.6\textwidth]{First_peak_airway_2.pdf}
% 	\caption{LEFT: An example of the posterior distribution. The maximum peak corresponds to the point of the airways where the cartilage no longer support the airways. TOP RIGHT: Reconstructed cross-sectional areas at 48mm before the maximum peak. The baseline and follow-up slice are displayed on the left and right respectively. BOTTOM RIGHT: Reconstructed cross-sectional areas at 54mm after the peak. The branches after the bifurcation are no longer supported by cartilage.}
% 	\label{First_peak_airway_2}
% \end{figure}

%To locate the region where the dilatation occurs, we used the posterior probability of the changepoint $p(\tau|\boldsymbol{y})$. We assume that IPF is a peripheral disease \cite{Jacob2015}. The dilatation of the airways begins at the distal point and extends towards the central part of the lung. We removed noise from the posterior by rescaling the distribution to 1 and setting values lower than 0.05 to zero. We obtain possible candidates for points of dilatation by considering local maximum values with a minimum peak distance of 20mm. Finally the candidate nearest to the distal point is considered to be the point $t$ at which the airway dilates.

For airways affected by IPF, dilatation starts at the distal point and progresses in the proximal direction \cite{Jacob2015}. Therefore, we topologically can assume that each affected airway undergoes a single changepoint from which dilatation starts. To locate such unique changepoint, we consider the posterior probability of the changepoint $p(\tau|\boldsymbol{y})$, and perform the following post-processing steps. 

% On each airway track, the proximal region are surrounded by cartilage \cite{Weibel1963}. As the airway track lose cartilage support, different biological noise occurs thus causing a changepoint. We eliminate such biological changepoint by setting the first 60mm of $p(\tau|\boldsymbol{y})$ to zero. Finally, we selected highest peak on the modified posterior $p(\tau|\boldsymbol{y})$ as the point of dilatation.

% (Ryu0
On each airway track, the proximal region is surrounded by cartilage \cite{Weibel1963}. As the airway track loses cartilage support, the geometry of the airway changes and results in a changepoint. Since such biological changepoint is independent of the disease state and occurs prior to dilatation, we eliminate it by discounting the most proximal peak in the posterior distribution $p(\tau|\boldsymbol{y})$. We then selected highest peak on the modified posterior $p(\tau|\boldsymbol{y})$ as the final estimation for the starting point of dilatation.

% For airways affected by IPF, traction bronchiectasis starts at the distal point and progress in a proximal direction \cite{Jacob2015}. Thus topologically, there is only one point of dilatation. To locate the point of dilatation, we consider the posterior probability of the changepoint $p(\tau|\boldsymbol{y})$, and perform several post processing steps.

% On each airway track, the proximal region are surrounded by cartilage \cite{Weibel1963}. As the airway track lose cartilage support, different biological noise occurs thus causing a changepoint. We eliminate the biological changepoint by setting the first 60mm of $p(\tau|\boldsymbol{y})$ to zero. Finally, we selected highest peak on the modified posterior $p(\tau|\boldsymbol{y})$ as the point of dilatation.

% To locate the point of dilatation we use the posterior probability of the changepoint $p(\tau|\boldsymbol{y})$ and input anatomical knowledge of the airways. For the first few branch generations, we assume the airway is supported by cartilage. The biological changes will have a different noise level between branches with and without cartilage support. Thus creating a changepoint on the arc length at end of the airways with the cartilage. We overcome this problem by setting the first 60mm of the posterior probability, $p(\tau|\boldsymbol{y})$. We then chose the point of dilatation as the maximum point on the modified posterior.

%%%%%%%%%%%%%%%%%%%%%%%%%%%%%%%%%%%%%%%%%%%%%%%%%%%%%%%%%%%%%%%%%%%%%%%%
\section{Evaluation \& Results}
%\comment{(Ryu): You need a short summary of evaluation you perform sec.~3.1 and 3.2. For example: ``in this section, we aim to evaluate the proposed method on datasets X, Y, Z. Firstly, we compare the accuracy of the method on dataset X with simulated geometric deformation of airways. Secondly, we assessed the utility of our method on a real dataset Y by comparing against the labels from radiologists."}

We evaluated our proposed method with two experiments. Firstly, we used a set of healthy airways, in which we had simulated dilatation to quantify the accuracy of our method and compared the results with conventional changepoint methodologies. Secondly, we assessed the clinical utility of our method on a dataset of airways affected by IPF. We compared our labels with those from two experienced thoracic radiologists. The properties of the images used in both experiments are displayed on Tab. \ref{long_make_and_voxel_size}.

\begin{table}
	\centering
	\begin{tabular}{ c | c | c | c | c | c }
		Experiment & Patients & BL Voxel Size & FU Voxel Size &  Airways & Time between scans \\ \hline
		1 & 1 & 0.67, 0.67, 1.00 & 0.56, 0.56, 1.00 & 6 & 9M 6D \\
		1 & 2 & 0.63, 0.63, 1.00 & 0.78, 0.78, 1.00 & 7 & 35M 6D \\
		1 & 3 & 0.72, 0.72, 1.00 & 0.72, 0.72, 1.00 & 1 & 5M 22D \\ \hline
		2 & 1 & 0.72, 0.72, 1.00 & 0.64, 0.64, 1.00 & 3 & 12M 5D \\
		2 & 2 & 0.67, 0.67, 1.00 & 0.87, 0.87, 1.00 & 1 & 10M 24D\\ 
		\hline
	\end{tabular}
	\vspace{3mm}
	\caption{Table of the image properties of voxel size, number of airways used and the time between scans. The airways were selected by a trained radiologist R1. All patients in Experiment 1 do not have IPF. Abbreviation: BL - Baseline Scan, FU - follow-up Scan, M - Months, D - Days.}
	\label{long_make_and_voxel_size}
\end{table}

% \begin{figure}
% 	\centering
% 	\includegraphics{Test_table_experminets.pdf}
% 	\caption{Table of the image properties of voxel size, number of airways used and the time between scans. The airways were selected by a trained radiologist R1. All patients in Experiment 1 do not have IPF. Abbreviation: BL - Baseline Scan, FU - follow-up Scan, M - Months, D - Days.}
% 	\label{long_make_and_voxel_size}
% \end{figure}

% \comment{(Ryu): Perhaps you could make a set-up section and chuck all the details of the datasets and the method configurations e.g. burn-in, number of iterations of RJMH, thresholding density, and max number of changepoints for the baseline methods, etc. I normally find this kind of structure useful as the messages of respective experiments would be clearer and if I wanted any more details about the datasets/methods, I would know where to refer to.}
\vspace{-10mm}
\subsection{Disease simulation}
% \comment{front load key finding or the aim of this experiment.}
% \ryu{the effects of IPF}{the airway dilatation caused by IPF}

To quantitatively assess accuracy, a ground truth is required. To this end, we applied our point detection algorithm on augmented healthy airway series to simulate the airway dilatation caused by IPF. After obtaining written informed consent from 3 patients, a trained radiologist (R1) selected 14 pairs of healthy airways in both baseline and follow-up scans. The image properties are displayed on Tab. \ref{long_make_and_voxel_size}. They were acquired from different scanners and used different reconstruction kernels. The airways were pre-processed as described in Sec. \ref{PreProMethod} to produce a function of area change along the length of the airway. We interpret this output signal as the overall noise caused from normal biological changes and acquisition.

\begin{figure}
	\centering
	\includegraphics[width=1.0\linewidth]{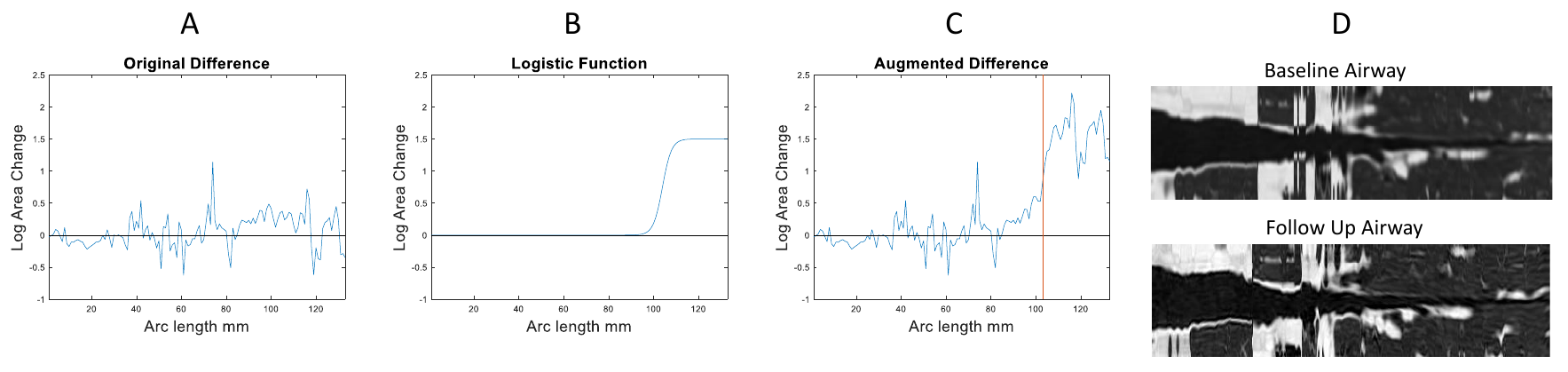}
	\caption{(A). An example of the area change of an airway remaining healthy on both baseline and follow-up scan. (B). A logistic function we constructed to simulate a dilatation due to IPF. (C). The augmented area change. The red line corresponds to our ground truth as the starting point of dilatation $\alpha$. (D). The corresponding reconstructed airways. The images were acquired using Quan et al \cite{Quan2018}.}
	\label{Method_aug}
\end{figure}

We modelled the change in dilatation with a logistic function;
\begin{equation}
    \boldsymbol{l} = \frac{M}{1+e^{-k(x - \alpha)}}.
\end{equation}
We interpreted $M$ as the magnitude of dilatation and $\alpha$ as the point of dilatation. The parameters $\alpha$ are set such that the dilatation starts 10-40mm from the distal point in increments of 5mm. In addition, we set $M$ to range from 0.3-3 in increments of 0.25. Finally we set $k = 0.5$mm$^{-1}$, in order to create an abnormal increase in cross sectional area.  To simulate the dilatation on the airway; the logistic function was added to the area change of the healthy airway, as shown on Fig. \ref{Method_aug}. We applied every permutation of $M$ and $\alpha$ on each of the 14 healthy airways. After the signal augmentation, we applied our proposed Bayesian changepoint detection algorithm.

%\textcolor{red}{Please just index them. e.g. (a), (b), (c), (d). Axis font/lines too small/thin. }

The proposed method was compared against two conventional methods. First, we use a basic thresholding method. For a given signal, we applied a 5mm moving mean average and thresholded the point at which the signal reached above the upper quartile from the right hand side. Secondly, we implemented the method based on Lavielle \cite{Lavielle2005}, in summary we consider $K$ changepoint and these changepoints $y_{i}$, minimize the function: 
\begin{equation}
J(K) = \sum^{K-1}_{r=0} \sum^{k_{r+1}-1}_{i = k_{r}} \Delta(y_{i},y_{k_{r}:k_{r+1}-1})  + \beta K, 
\end{equation}
where $\beta$ is modified such that the function finds less than $K$ changepoints. For our paper, different $\Delta$ were evaluated and we found 
\begin{equation}
    \Delta(y_{i},y_{k_{r}:k_{r-1}-1})= y_{i} - \meano(y_{k_{r}:k_{r+1}-1})
\end{equation}
gives the most accurate results. To replicate the post processing of our proposed method, we consider $K = 2$ possible changepoints. This takes into account the changepoint caused by the support cartilage. Finally, we set a minimum distance of 20mm. Once we acquired the changepoints, the most peripheral point was chosen as the point of dilatation. The implementation was performed through Matlab inbuilt function; \texttt{findchangepts}\footnote{\url{https://www.mathworks.com/help/signal/ref/findchangepts.html} last accessed on \today.}. 

%To replicate the post processing of our proposed method, we set consider all possible changepoints with a minimum distance of 20mm and chose the most peripheral point as the point of dilatation.

% For comparison of the methodology; using the same augmented data, \ryu{we assessed the performance of changepoint detection based on maximizing the log likelihood}{the likelihood is never defined for this model in the method section. It is a different model to the one we use. Also, how do methods (b), (c), (d), (e) differ? You need to give a brief description of them and explain why they are meaningful baselines.}.  The implementation was performed using the $findchangepts$ function from Matlab\footnote{\url{https://www.mathworks.com/help/signal/ref/findchangepts.html} last accessed on \today.}. We tested all statistics options on $findchangepts$: mean, root mean square, standard deviation and both mean and slope. For every statistic option, we set the maximum number of changes as 30 and the minimum distance of 20mm. We choose the point of dilatation to be the most peripheral changepoint. Finally, a basic thresholding method was implemented. For a given the signal, we applied a 5mm moving mean average and threshold the point where signal reaches above the upper quartile from the right hand side.
%Each combination of $M$ and $\alpha$ was applied to all 14 airway pairs. The metric for comparison is the error difference $e = t - \alpha$. %Figure \ref{Results_2_main}, shows a decrease in accuracy correlates to a smaller dilatation and increase in proximal start of dilatation. [DISCUSS PRECISION]

\begin{figure}
	\centering
	\includegraphics[trim={0 10 0 0},clip,width=\textwidth]{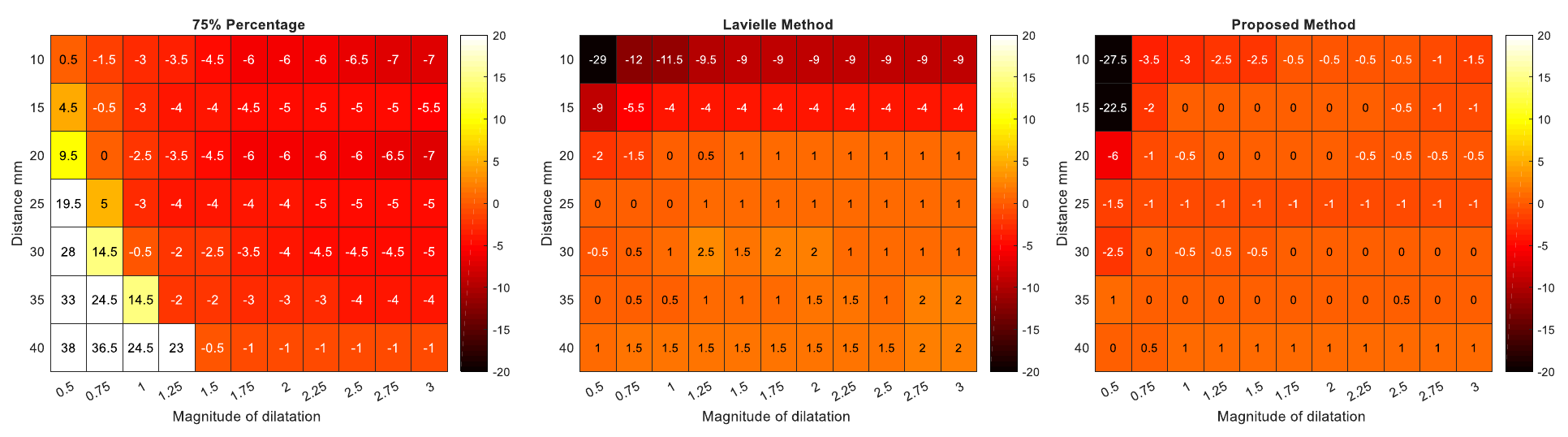}
	\caption{Heatmap showing the accuracy of each method. (Left). Thresholding method. (Middle). Method from Lavielle \cite{Lavielle2005}. (Right). Our proposed method. The colour scale is the same on all of the heatmaps.}
	\label{Overleft_method_acc_conparsion}
\end{figure}

%A: Thresholding method, B-E: Results from maximizing the log likelihood using the respective statistics. F: Our proposed method. Each entry corresponds to the medium average displacement error over 14 airways. The colour scale is the same on all of the heatmaps.

Fig.~\ref{Overleft_method_acc_conparsion} shows the accuracy for each individual method as a heatmap. The metric used to quantify accuracy was the displacement between the ground truth and the changepoint given by the proposed method. A positive displacement (mm) corresponds to an overestimation of the ground truth towards the distal point. Each entry on the heatmap corresponds to the median average of all displacements over 14 airway pairs.

%The proposed method is accurate within 3mm in locating the point of dilatation independent of the magnitude of the dilatation for peripheral distances, 10-30mm \ryu{}{as long as it is above $0.75$. This is is good because}. \comment{ (Ryu): I think there are a lot more we can talk about here. The accuracy of other baseline methods are never mentioned. You need to describe the error patterns and why they differ. Need to explain why the error pattern of your method is more desirable than the other ones. }

When the magnitude of dilation is larger than $M > 0.75$, our proposed method achieves consistently higher accuracy than Lavielle \cite{Lavielle2005}. We note that the accuracy gain in the peripheral regions of the airways at $\alpha=$ 10-30mm from the distal point are the most clinically relevant in IPF as lung parenchymal damage begins in the lung periphery and progresses proximally \cite{Jacob2015}. Furthermore on the same peripheral regions $\alpha=$ 10-30mm, the baseline method showed systematic bias in accuracy towards the central airways. This was due to the baseline method being influenced by outliers from the longer expanses of normal airway regions. The proposed method uses the t-distribution as the likelihood thus making it robust to possible outliers within the data \cite{Prince2012}. On the other hand, the proposed method suffers from poor accuracy below magnitudes of dilatation $M = 0.75$. However, in physical terms a dilatation of $M = 0.75$ corresponds to a percentage increase in cross sectional area of $e^{0.75} - 1 \approx 112\%$. This is within the range of normal biological change of the airways \cite{Gazourian2017}.

\subsection{Application to Airways affected by IPF}
% \comment{(Ryu): As previous comment: front load key finding or motivation. What is the main take away or whay is the main thing that you want to test here? Describe that in a couple of sentences so the reader knows immediately what you are trying to do.}
% \ryu{We applied our proposed methods on airways affected by IPF on longitudinal CT scans}{Which data set you are talking about now? If different from the first, you need to make this clearer. I would strongly encourage you to make a section solely on datasets and method configurations to improve the clarity}.

We applied our method to airways affected by IPF. The purpose was to compare our measurement to the labels provided by a radiologist and compute the volume change of the identified diseased airway regions. For our dataset, we acquired 4 airway pairs from 2 patients after obtained a waiver for consent from the local Research Ethics Committee. All airways were judged by the radiologist R1 to be dilated as a consequence of IPF on baseline and to have visually worsened on follow-up imaging. Image properties are displayed in Tab.~\ref{long_make_and_voxel_size}.

\begin{figure}
    \centering
    \includegraphics[trim={0 120 0 120},clip,width=\textwidth]{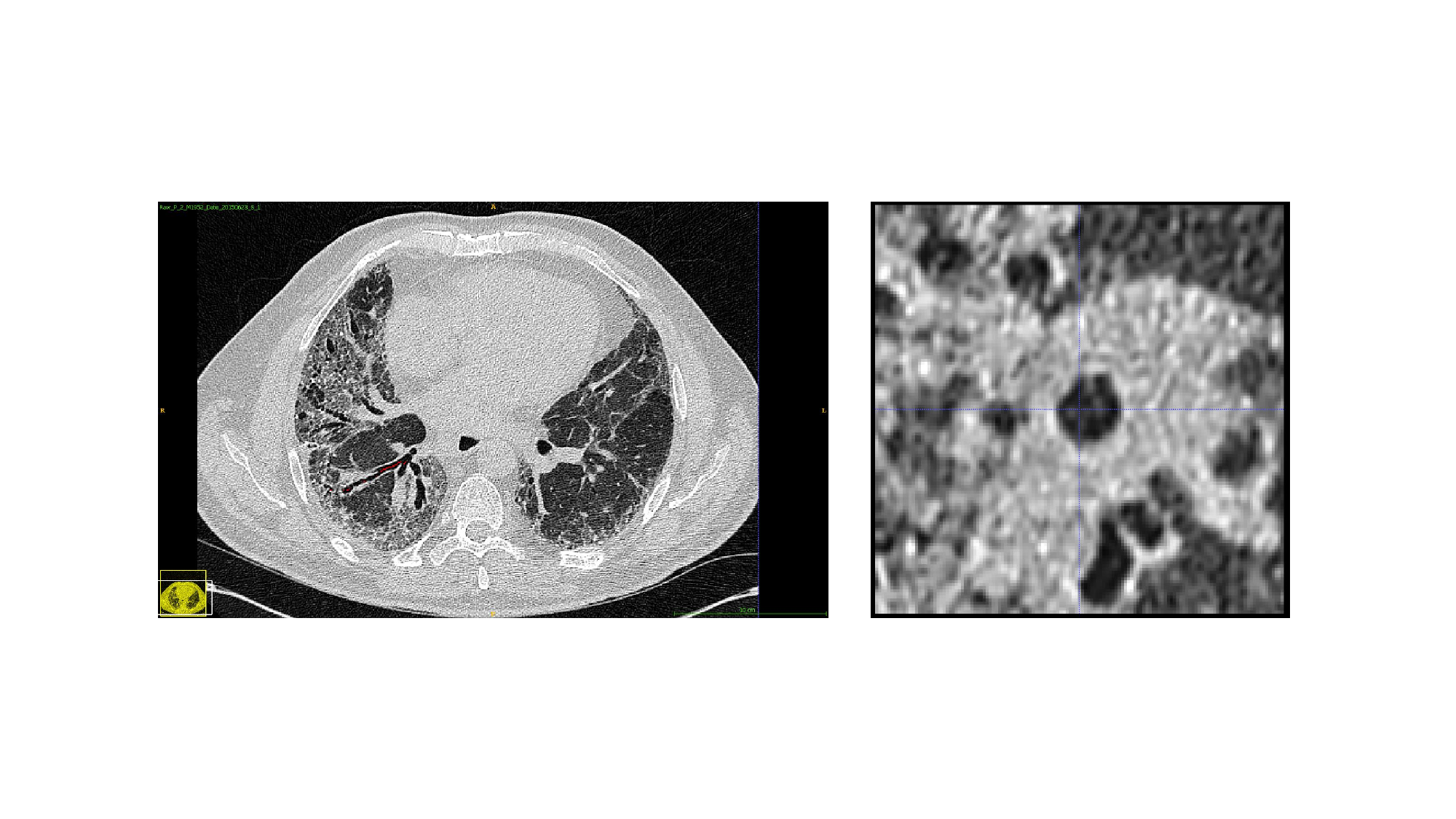}
    \caption{Example views of images in two measurement protocols employed by radiologists to locate starting points of dilatation. (Left). Protocol 1 based on the axial slice. (Right). Protocol 2 based on the reconstructed cross sectional planes.}
    \label{Vis_radio_methods}
    
\end{figure}

We compared the performance of our method against two trained thoracic subspecialist radiologists. Two radiologists R1, R2 identified the point of at which a given airway was seen to demonstrate increased dilatation on the follow-up CT scan. To assessed the reproducibility of manual labeling, each radiologist labelled the same airway twice through two different protocols. In the first method, the radiologists interrogated axial CT images. Using 2 separate workstations and the airway centreline, the radiologists identified the point on the centreline (on the follow-up scan) where the airway demonstrated definitive worsened dilatation. For the second method, the radiologist compared the aligned reconstructed cross-sectional planes on baseline and follow-up scans. The radiologist then selected the slice where the airway had worsened when evaluated against the baseline scan. An example of both protocols are displayed on Fig.  \ref{Vis_radio_methods}.

% The results of the labelling are displayed in Figure \ref{Final_plot_rad_vs_porposed}. It shows some interobserver variation on labelling using the axial slice. In addition, there is interobserver variation when comparing different visualization method. %\comment{(Ryu): we should try to explain the differences between our method and the radiologists in Airway 2 and 3 for which we constantly over-estimate in Fig.~\ref{Final_plot_rad_vs_porposed}. Is our method actually failing here? or do you think the radiologists may be giving wrong labels confounding dilatation with noise? }

\begin{figure}[h]
	\centering
	\includegraphics[trim={0 0 0 0},clip,width=\textwidth]{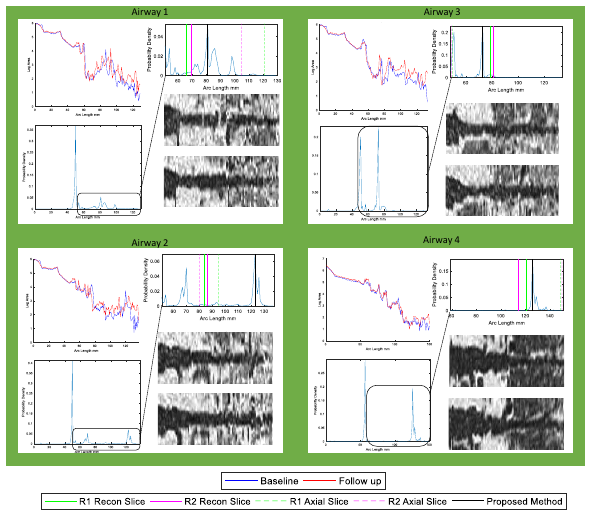}
	\caption{The log cross sectional area and posterior distribution $p(\tau|\boldsymbol{y})$ for each of the four airway pairs. Airways 1,2.3 come form the same patient. In the magnified region (black) we compared the labels from our proposed method with the radiologist. Furthermore, we displayed the reconstructed slices within the magnified region.}
	\label{Final_plot_rad_vs_porposed}
\end{figure}

\begin{figure}[h]
	\centering
	\includegraphics[trim={0 0 0 0},clip,width=1.0\textwidth]{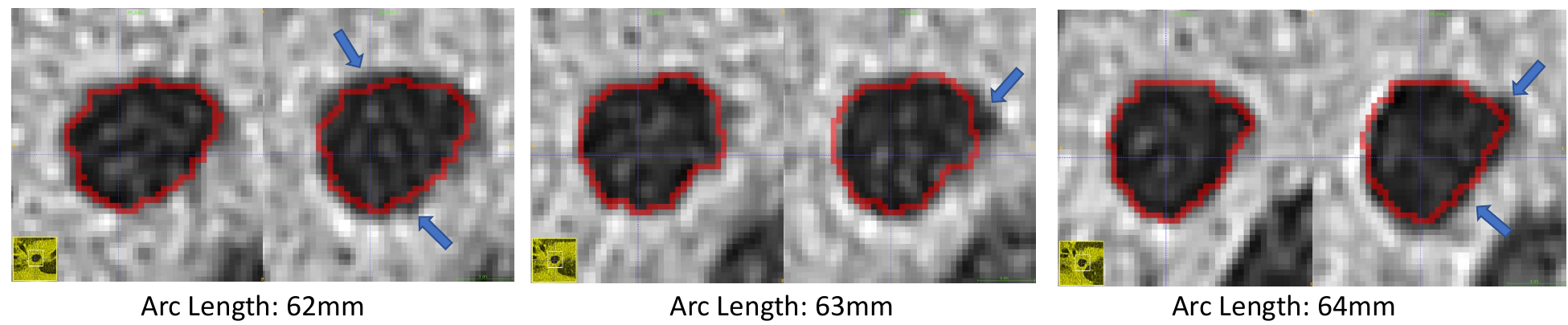}
	\caption{A row of three consecutive reconstructed slices in Airway 2 located on the arc length of 62-64mm. Each slice shows the airway lumen at baseline (left) and follow-up (right). The boundary delineation (red) from the baseline are superimposed on the follow-up scan. The blue arrows indicate pixels from the lumen outside the boundary.}
	\label{zoom_Airway_2}
\end{figure}

Fig.  \ref{Final_plot_rad_vs_porposed} compares the predictions of our method with the labels from radiologists obtained in two different protocols. The results indicate that the predictions for Airway 1, 3 and 4 are within the range of the radiologists' labels. In the case of Airway 2, although our method based on the maximum peak overestimates with respect to the radiologists' predictions, the posterior distribution contains another equally probable peak that underestimates the radiologists' labels (see the second highest peak at 70mm), potentially indicating a more proximal point of dilatation. To test this, we delineated the boundary of the lumen on the reconstructed cross sectional slices at baseline in the neighbouhood of this peak, and Fig.~\ref{zoom_Airway_2} shows the initial few slices (62-64mm). We observed that, when the delineated boundary from baseline was superimposed on the follow-up scan, the boundary is contained inside of the follow-up lumen with several lumen pixels are consistently outside from the boundary. Thus, this result indicates that the starting point of dilatation is more proximal than the labels from the radiologists.
% \vspace{-10mm}
% lol

%It shows that the labels from our proposed method in Airway 1 and 4 are within range of the labels from the radiologists. On the other hand, the proposed method shows a systematic overestimation in both Airway 2 and 3, this is possibly due to the ad hoc post-processing of the posterior distribution. However, if we consider the entire posterior distribution which contains multiple changepoints. For Airway 3, labels in particular from cross sectional protocol lies on higher densities regions in the posterior distribution. For Airway 2, the labels  %We remark that the labels are in between the larger density parts of the posterior distribution.

To demonstrate the clinical utility of locating the starting point of airway dilatation, we compared longitudinal airway volume changes in diseased and healthy regions of each of the airway tracks. To find the volume of the airway, we considered the aligned signals,  $f_{F}$,$f_{B}$. These signals are measurements of area against the airway arc length. Thus volume can be computed via the area under the curve. For our work, we used the trapezium rule to find the volume. Three volumetric regions were considered: (i) the entire airway $V_{c \rightarrow d }$ i.e. from the carina, $c$ to the distal point, $d$, (ii) carina to the dilatation point $t$, denoted as $V_{c \rightarrow t }$. (iii) the dilatation point to the distal point, denoted by $V_{t \rightarrow d }$. Note that $V_{c \rightarrow d } = V_{c \rightarrow t } \cup V_{t \rightarrow d }$ and $V_{c \rightarrow t }$ does not overlap with $V_{t \rightarrow d }$. Tab.~\ref{vol_tdis}, shows the results of the percentage volume change. The volume change in $V_{t \rightarrow d}$ had greater sensitivity for selecting progressive airway dilatation in IPF than the volume change in the entire airway $V_{c \rightarrow d }$.

\begin{table}
	\centering
	
	\begin{tabular}{| c | c | c | c  |}
	    \hline 
		Airway & PVC of $V_{c \rightarrow d }$ & PVC of $V_{t \rightarrow d }$ & PVC of $V_{c \rightarrow t}$ \\ \hline
        1 & 2.6 \%  & 32.9\%   & 1.1\%  \\
        2 & 3.2 \%  & 129.7\%  & 2.2\% \\
        3 & 2.6 \%  & 47.4 \%  & -0.3\%  \\
        4 & 7.4 \%  & 48.4 \%  & 7.1\% \\\hline 
	\end{tabular}
	\vspace{3mm}
	\caption{The percentage volume change (PVC $\%$) for each region of the airway.}
	\label{vol_tdis}
\end{table}

% \begin{figure}[h]
% 	\centering
% 	\includegraphics{Test_table.pdf}
% 	\caption{The percentage volume change (PVC) for each specified regions of the airway track.}
% 	\label{vol_tdis}
% \end{figure}

\section{Discussion \& Conclusion}
In this paper, we modelled changes in area along airway tracks as a time series with the purpose of detecting abnormal dilatation caused by IPF using Bayesian changepoint detection. Experiments on simulated data show that our model is able to detect the starting location of airway dilatation with superior accuracy than the relevant baseline methods. The results on the IPF longitudinal dataset display reasonable agreement with radiologists, while in one case indicating a more plausible location of dilatation, potentially missed by the experts.

Identifying changepoints and thereby calculating a change in airway dilatation over time could become a sensitive measure of IPF aggravation. This would form an important secondary endpoint in drug trials, where our measurements could indicate whether a new drug ameliorates disease progression better than existing medications. In the future, we hope to evaluate such utility of our proposed method on a larger cohort of IPF subjects. Furthermore, our method is applicable to other lung airway diseases, characterised by dilatation (such as cystic fibrosis) or other geometrical deformations, and such extensions remain valuable future work. 

\section{Acknowledgements}

Kin Quan is supported by the EPSRC-funded UCL Centre for Doctoral Training in Medical Imaging (EP/L016478/1) and the Department of Health NIHR-funded Biomedical Research Centre at University College London Hospitals. Ryutaro Tanno is supported by Microsoft Research Scholarship. Micheal Duong is funded by the Centre for Doctoral Training in Financial Computing and Analytics. Arjun Nair is funded by National Institute for Health Research Biomedical Research Centre. Mark Jones and Christopher Brereton are supported by National Institute for Health Research. Joseph Jacob is a recipient of Wellcome Trust Clinical Research Career Development Fellowship 209553/Z/17/Z.

An adapted version of this manuscript was accepted to The 10th International Workshop on Machine Learning in Medical Imaging (MLMI 2019). In conjunction with MICCAI 2019, Shenzhen, China. It is cited as follows: Quan K. et al. (2019) Modelling Airway Geometry as Stock Market Data Using Bayesian Changepoint Detection. In: Suk HI., Liu M., Yan P., Lian C. (eds) Machine Learning in Medical Imaging. MLMI 2019. Lecture Notes in Computer Science, vol 11861. Springer, Cham. DOI: \url{https://doi.org/10.1007/978-3-030-32692-0_40}.

\bibliographystyle{plain}
\bibliography{mybibliography}

\end{document}